\documentclass[numsec,webpdf,modern,large,namedate]{oup-authoring-template}

\graphicspath{{Fig/}}

\usepackage{doi}
\usepackage{pdfpages}

\theoremstyle{thmstyleone}%
\theoremstyle{thmstyletwo}%
\theoremstyle{thmstylethree}%

\begin{document}

\journaltitle{arXiv preprint}
\DOI{DOI HERE}
\copyrightyear{2023}
\pubyear{2023}
\appnotes{Applications Note}

\firstpage{1}

\subtitle{Data and text mining}

\title[survex]{survex: an R package for explaining machine learning survival models}

\author[1]{Mikołaj Spytek}
\author[1]{Mateusz Krzyziński}
\author[2,3]{Sophie Hanna Langbein}
\author[1,4]{Hubert Baniecki}
\author[2,3,5]{Marvin N. Wright}
\author[1,4$\ast$]{Przemysław Biecek}

\authormark{M. Spytek et al.}

\address[1]{\orgdiv{MI2.AI}, \orgname{Warsaw University of Technology},\orgaddress{
\country{Poland}}}
\address[2]{\orgname{Leibniz Institute for Prevention Research and Epidemiology – BIPS},\orgaddress{
\country{Germany}}}
\address[3]{\orgname{Faculty of Mathematics and Computer Science, University of Bremen},\orgaddress{
\country{Germany}}}
\address[4]{\orgdiv{MI2.AI}, \orgname{University of Warsaw},\orgaddress{
\country{Poland}}}
\address[5]{\orgname{Section of Biostatistics, Department of Public Health, University of Copenhagen},\orgaddress{
\country{Denmark}}}

\corresp[$\ast$]{Corresponding author. \href{email:email-id.com}{przemyslaw.biecek@pw.edu.pl}}

\received{Date}{0}{Year}
\revised{Date}{0}{Year}
\accepted{Date}{0}{Year}


\abstract{
\textbf{Summary:} Due to their flexibility and superior performance, machine learning models frequently complement and outperform traditional statistical survival models. However, their widespread adoption is hindered by a lack of user-friendly tools to explain their internal operations and prediction rationales. To tackle this issue, we introduce the survex R package, which provides a cohesive framework for explaining any survival model by applying explainable artificial intelligence techniques. The capabilities of the proposed software encompass understanding and diagnosing survival models, which can lead to their improvement. By revealing insights into the decision-making process, such as variable effects and importances, survex enables the assessment of model reliability and the detection of biases. Thus, transparency and responsibility may be promoted in sensitive areas, such as biomedical research and healthcare applications. 
 \\
\\
\textbf{Availability and Implementation:} survex is available under the GPL3 public license at \href{https://github.com/modeloriented/survex}{https://github.com/modeloriented/survex} and on CRAN with documentation available at \href{https://modeloriented.github.io/survex/}{https://modeloriented.github.io/survex}.\\
\textbf{Contact:} \href{przemyslaw.biecek@pw.edu.pl}{przemyslaw.biecek@pw.edu.pl} \\
}


\maketitle

\section{Introduction}
Survival analysis focuses on the estimation of time-to-event distributions while considering the effects of censoring. Over time, this field has witnessed substantial progress, initially driven by conventional statistical approaches like the Cox proportional hazards model \citep{cph}. However, inherent limitations and challenges within this domain have spurred the integration of machine learning methodologies, introducing enhanced performance and flexibility \citep{survml-survey}. This convergence of statistics and machine learning is particularly evident in the R programming language \citep{R}, where various frameworks have been developed to facilitate the application of both traditional statistical models and modern machine learning techniques for survival analysis tasks. 

Despite the promising potential of integrating machine learning into biomedical research and healthcare, the opaque nature of black-box models has raised valid concerns \citep{iml-healthcare}. In response, interpretable machine learning and explainable artificial intelligence methods has emerged as a viable solution \citep{ema, iml-molnar}, including software packages in R \citep{dalex,iml}. However, these software packages cannot handle censoring and do not provide explanations for survival models. To fill this gap, we propose \textbf{survex} as an innovative solution that provides comprehensive explanations for entire models and individual predictions, alongside performance measures and a unified prediction interface. Operating within the R environment, \textbf{survex} supports numerous packages with survival models while maintaining flexibility to integrate others. By integrating explanations into the modeling and analysis process, \textbf{survex} aspires to empower stakeholders, particularly in critical domains like healthcare, with a deeper understanding of the model's predictions and underlying rationales, ultimately promoting trust and informed decision-making.

\begin{figure*}
    \centering
    \includegraphics[width=0.95\textwidth]{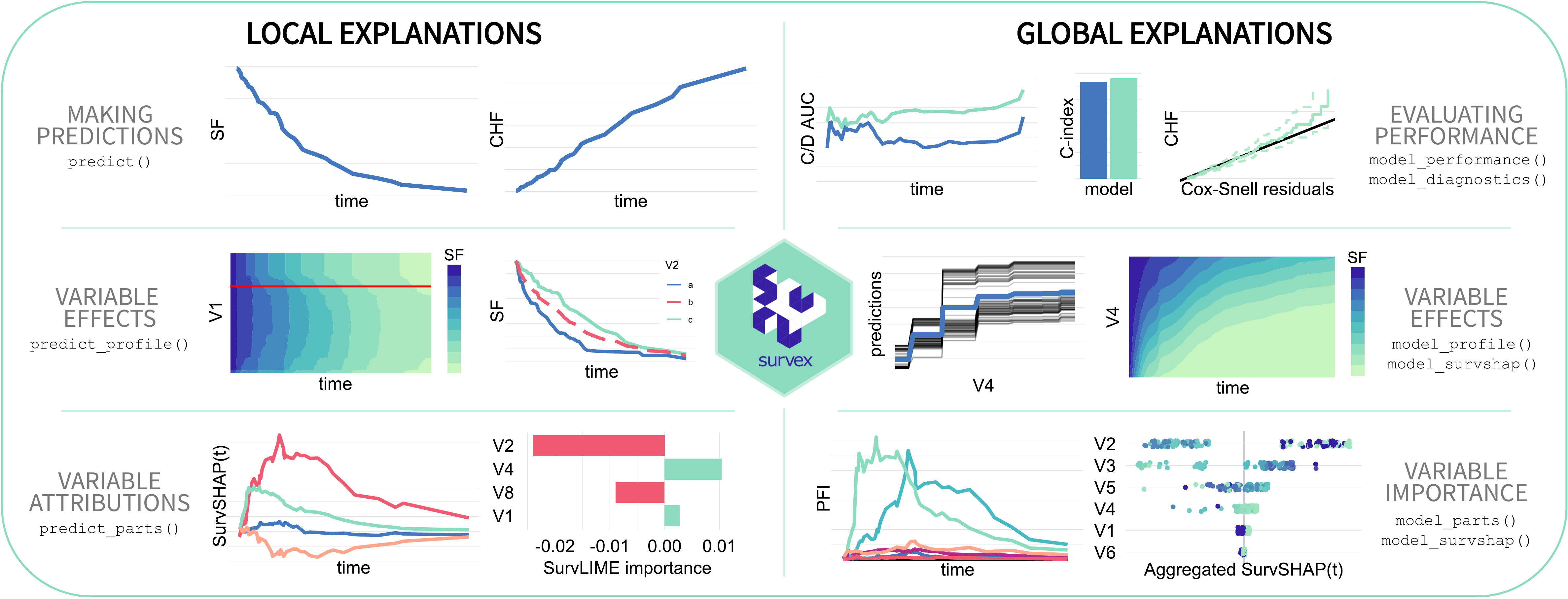}
    \caption{Explanations and functionalities available in the \textbf{survex} package. The methods are divided into local (concerning individual predictions) and global (concerning the model). The diagram illustrates simplified examples of the visualizations of selected explanations in each category. A complete list of functionalities with documentation is available at \url{https://modeloriented.github.io/survex}.}
    \label{fig:survex}
\end{figure*}

\section{Related work}
In the domain of survival analysis, various software packages offer a range of statistical and machine learning methods. In R, an essential component is the \textbf{survival} package \citep{survival}, which contains the fundamental statistical models like the Cox proportional hazards \citep{cph} and accelerated failure time \citep{aft} models. Alongside this, \textbf{randomForestSRC} \citep{rfsrc} and \textbf{ranger} \citep{ranger} packages provide implementations of the notable random survival forest algorithm \citep{rsf}. Complementing these, the \textbf{survivalmodels} package \citep{survivalmodels} offers deep neural networks within the survival analysis paradigm. To streamline consistency within this diverse toolkit, frameworks such as \textbf{mlr3proba} \citep{mlr3proba} and \textbf{censored} \citep{censored} extend the \textbf{mlr3} \citep{mlr3} and \textbf{parsnip} \citep{parsnip} frameworks, respectively, to provide standardized ways of using different survival models. 

To illustrate the wide range of model explainability tools available in the R environment, it is pertinent to highlight packages such as \textbf{DALEX} \citep{dalex} and \textbf{iml} \citep{iml}, which offer a diverse spectrum of XAI techniques. While there are many packages in this field, their core focus remains rooted in the domain of explaining classification and regression models. Adapting certain explanatory methods for survival models can be accomplished, but it requires careful adjustment due to the unique predictive nature of these models. To address this challenge, \textbf{survex} offers specifically tailored explanations that incorporate the time dimension inherent in the survival models' predictions. Furthermore, methods dedicated to explain survival models have been formulated, such as SurvLIME \citep{survlime} with a Python implementation \citep{pachon2023survlimepy} or SurvSHAP(t) \citep{survshap}. The \textbf{survex} package is equipped to incorporate also these advanced techniques, further enhancing its functionalities range. It is worth acknowledging the inspiration drawn from the \textbf{survxai} package \citep{survxai}, which has significantly influenced the development of \textbf{survex}. However, \textbf{survex} offers a broader spectrum of functionalities, recently proposed explanation methods, and supports a wider range of models. In addition, \textbf{survxai} is no longer maintained.

\section{Implementation and functionalities}

The following section presents a brief overview of the functionalities of \textbf{survex}. More details about the explanation methods and implementation can be found in the accompanying \textbf{Supplementary Information} and package documentation. 

The design of \textbf{survex} draws inspiration from the \textbf{DALEX} R package \citep{dalex}, serving as an extension tailored to explain survival models. This is exemplified in a methodically structured interface, aligned with the structure proposed by \citet{iema}. The package facilitates the interpretation of survival models through its diverse functionalities, categorized into local and global contexts, as depicted in Figure~\ref{fig:survex}. \textbf{survex} is engineered as a model-agnostic framework adaptable to any model that returns predictions in the form of a survival function or cumulative hazards function.

The model-agnostic approach is implemented by the central component of the package -- the \texttt{explainer} object. Serving as a wrapper for survival models, it unifies their prediction interfaces and stores essential background data necessary for obtaining predictions and explanations. For models from widely-used packages, the \texttt{explainer} can be created automatically, by providing the model object to the \texttt{explain()} function. However, any model not adapted automatically by \textbf{survex} can be explained by  specifying the way of predicting the survival function. For the wrapped model, the unified prediction interface can be used by the \texttt{predict()} function, capable of generating predictions in the~form of survival function, cumulative hazard function, or relative risk. 

\vspace{-0.5em}
\subsection{Explanation methods}

Global explanations, concerning the whole model (dataset level), are marked with the \texttt{model} prefix, while local explanations, referring to individual predictions (observation level), are denoted by the \texttt{predict} prefix. Computing an explanation involves invoking the relevant function with the \texttt{explainer} object as the primary argument, supplemented by additional details. One of the key parameters is the \texttt{output\_type} -- the default selection of \texttt{`survival'} generates time-dependent explanations based on a survival function. However, it is also possible to select \texttt{`chf'} for explanations related to cumulative hazard function or \texttt{`risk'}, which results in more standard explanations based on a prediction in the form of relative risk (single number). 

The \texttt{model\_parts()} function outputs variable importance scores for the model. It leverages permutation variable importance \citep{breiman2001rf, pfi}, i.e., quantifies the extent by which performance metric values are impacted upon permuting values of a chosen predictor. 

Performance measures can also be used with the \texttt{model\_performance()} function, allowing users to comprehensively evaluate the models' predictive capabilities. This function also offers the possibility to prepare ROC curves at different time points, by treating the survival probability at a selected time point as the response for the classification task.

Furthermore, the \texttt{model\_diagnostics()} function facilitates diagnostic assessments through analysis of the residuals. It supports the calculation and visualization of martingale residuals, deviance residuals \citep{martingale_deviance}, and Cox-Snell residuals \citep{coxsnell}. 

Explanations obtained by the \texttt{model\_profile()} function reveal the influence of a specific variable on the model's predictions. They are constructed based on one of two distinct methodologies: partial dependence plots \citep{pdp} or accumulated local effects \citep{ale}. Moreover, for insight into potential interaction effects, \textbf{survex} offers the \texttt{model\_profile\_2d()} function, which generates profiles for two variables. 

Using the \texttt{predict\_parts()} function results in explanations that reveal the contributions of variables to a model's prediction for a selected observation. In \textbf{survex}, these insights can be obtained using one of the two methodologies. The default method is SurvSHAP(t) \citep{survshap}, leveraging SHAP values applied to the survival function to give the variable attributions at different times, alongside their aggregations over time. Alternatively, the SurvLIME approach fits a surrogate Cox model in the local neighborhood of the selected observation and uses its coefficients as the explanation.

The \texttt{predict\_profile()} function is related to explanations concerning a single variable's impact on a specific prediction. These insights are derived via the individual conditional expectation method \citep{ice}, also known as the \textit{ceteris paribus} method, as it involves altering the values of a single variable while keeping all others constant. These results can be analyzed together with partial dependence plots, which are their average.

The \texttt{model\_survshap()} function streamlines the process of accessing SurvSHAP(t) explanations for a specified set of observations. Beyond individual explanations, this function aggregates SurvSHAP(t) values, revealing global insights into the model's behavior. Moreover, it incorporates accessible visualization methods, including SurvSHAP(t) bee swarm and dependence plots, inspired by the well-established \textbf{shap} Python package \citep{shap}.

It should be noted that methods using permute-and-predict mechanism have been criticized for producing misleading results when dealing with strongly correlated variables \citep{Hooker2019UnrestrictedPF}, and there are alternative methods specifically designed to address these challenges \citep{Delicado2019UnderstandingCP}. Thus, techniques like permutation variable importance, individual conditional expectation and partial dependence plots should be used cautiously.

\vspace{-0.5em}
\subsection{Visualizations}
Within \textbf{survex}, various visualizations accompany its explanations. Plots are prepared using the \textbf{ggplot2} package \citep{ggplot2} through the implemented \texttt{plot()} function called on the object returned by the explanation. Extensive user customization is enabled by adjustable parameters within the plotting functions, augmented with advanced functions available in \textbf{ggplot2}. Plots can be created for multiple \texttt{explainer} objects at once, allowing the user to compare and differentiate the explanations for different models and observations.

\vspace{-0.5em}
\subsection{Applications} 
\textbf{survex} has already demonstrated its applicability in the field of biomedical research and healthcare. \citet{chen2023machine} employed the package to find out the relative importance of variables in survival models predicting sporadic pancreatic cancer. \citet{nachit2023ai} used it to analyze partial dependence plots of different body composition parameters extracted from computer tomography scans in a random survival forest. Additionally, we successfully applied \textbf{survex} to explain model bias in predicting hospital length of stay~\citep{baniecki2023hospital}.

\vspace{-0.5em}
\section{Future work}
Currently, \textbf{survex} handles the most common case of right-censored data with a single type of event. However, a future roadmap envisions its extension to cover alternative censoring types and accommodate competing risk models. Moreover, \textbf{survex} can be easily extended with additional explanation techniques, e.g., counterfactual explanations proposed by \citet{counterfactuals}, or more adaptations of existing methods known from classification and regression tasks.

\vspace{-0.5em}
\section*{Supplementary data}
Supplementary data are available at \textit{Bioinformatics} online.

\vspace{-0.5em}

\section*{Funding}
This work was supported by the National Science Centre [SONATA BIS 9 grant 2019/34/E/ST6/00052]; the Polish National Centre for Research and Development [INFOSTRATEG-I/0022/2021-00]; and the German Research Foundation (DFG) [Grants 437611051, 459360854]. 

\bibliographystyle{abbrvnat}
\bibliography{reference}

\includepdf[pages=-]{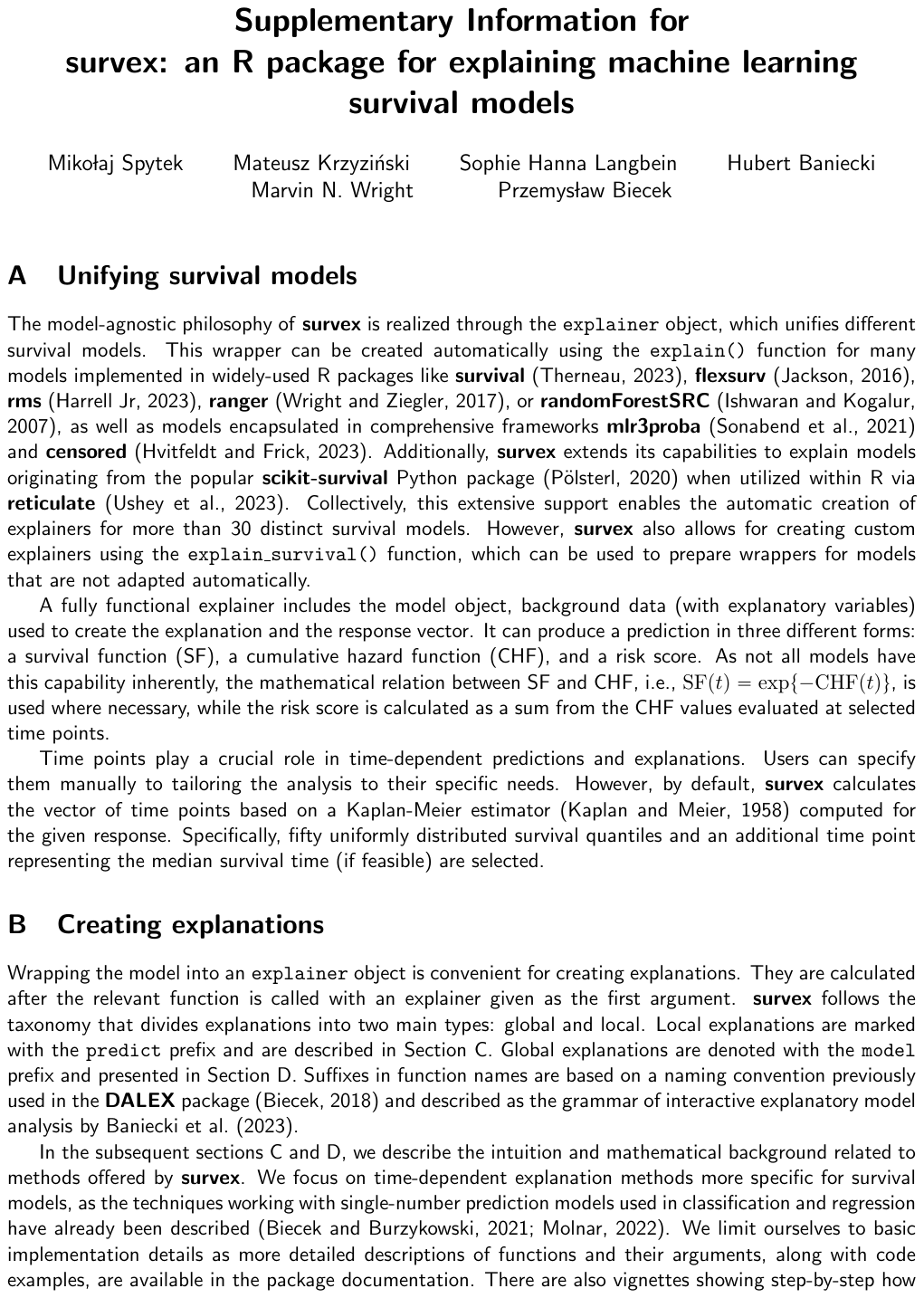}

\end{document}


\maketitle

\thispagestyle{empty}

\appendix
\section{Unifying survival models}

The model-agnostic philosophy of \textbf{survex} is realized through the \texttt{explainer} object, which unifies different survival models. This wrapper can be created automatically using the \texttt{explain()} function for many models implemented in widely-used R packages like \textbf{survival} \citep{survival}, \textbf{flexsurv} \citep{flexsurv}, \textbf{rms} \citep{rms}, \textbf{ranger} \citep{ranger}, or \textbf{randomForestSRC} \citep{rfsrc}, as well as models encapsulated in comprehensive frameworks \textbf{mlr3proba} \citep{mlr3proba} and \textbf{censored} \citep{censored}. Additionally, \textbf{survex} extends its capabilities to explain models originating from the popular \textbf{scikit-survival} Python package \citep{sksurv} when utilized within R via \textbf{reticulate} \citep{reticulate}. Collectively, this extensive support enables the automatic creation of explainers for more than 30 distinct survival models. However, \textbf{survex} also allows for creating custom explainers using the \texttt{explain\_survival()} function, which can be used to prepare wrappers for models that are not adapted automatically. 

A fully functional explainer includes the model object, background data (with explanatory variables) used to create the explanation and the response vector. It can produce a prediction in three different forms: a survival function (SF), a cumulative hazard function (CHF), and a risk score. As not all models have this capability inherently, the mathematical relation between SF and CHF, i.e., $\mathrm{SF}(t) = \exp\{-\mathrm{CHF}(t)\}$, is used where necessary, while the risk score is calculated as a sum from the CHF values evaluated at selected time points.

Time points play a crucial role in time-dependent predictions and explanations. Users can specify them manually to tailoring the analysis to their specific needs. However, by default, \textbf{survex} calculates the vector of time points based on a Kaplan-Meier estimator \citep{kaplan-meier} computed for the given response. Specifically, fifty uniformly distributed survival quantiles and an additional time point representing the median survival time (if feasible) are selected.

\section{Creating explanations}
Wrapping the model into an \texttt{explainer} object is convenient for creating explanations. They are calculated after the relevant function is called with an explainer given as the first argument. \textbf{survex} follows the taxonomy that divides explanations into two main types: global and local. Local explanations are marked with the \texttt{predict} prefix and are described in Section~\ref{sec:local}. Global explanations are denoted with the \texttt{model} prefix and presented in Section~\ref{sec:global}. Suffixes in function names are based on a naming convention previously used in the \textbf{DALEX} package \citep{dalex} and described as the grammar of interactive explanatory model analysis by \citet{iema}.

In the subsequent sections \ref{sec:local} and \ref{sec:global}, we describe the intuition and mathematical background related to methods offered by \textbf{survex}. We focus on time-dependent explanation methods more specific for survival models, as the techniques working with single-number prediction models used in classification and regression have already been described \citep{ema, iml-molnar}. We limit ourselves to basic implementation details as more detailed descriptions of functions and their arguments, along with code examples, are available in the package documentation. There are also vignettes showing step-by-step how to use the tool for specific use cases. Documentation, which is continuously updated, is available at \url{https://modeloriented.github.io/survex/}.

In mathematical description, we adopt the following notation. We use $f_t(\bx)$ to denote a prediction for an observation $\bx = [\bx^1, \dots, \bx^p]$ at time point $t$. The function $f$ is estimated by the model of interest, and $t$ ranges from $t_0$ to $t_k$ (the vector of time points contained in the explainer). Unless otherwise stated, $f$ can represent both survival function and cumulative hazard function, as the formulation of the methods is correct in both cases. When a differentiation is required, we use $f_t^\mathrm{SF}(\bx)$ for the survival function and $f_t^\mathrm{CHF}(\bx)$ for the cumulative hazard function. Moreover, we denote by $\bX$ the whole background dataset consisting of $n$ observations $\bx_1, \bx_2, \ldots, \bx_n$, each having $p$~variables. Further, by $\by = [(y_1, \delta_1), (y_2, \delta_2), \ldots, (y_n, \delta_n)]$, we denote the response -- survival target (pairs of observed times and event indicators) for this dataset.

\section{Local explanation methods}
\label{sec:local}

\subsection{Variable attributions}

Variable attribution methods serve to indicate how different variables contribute to the prediction obtained from the model for the selected observation. \textbf{survex} offers two different methods SurvSHAP(t) and SurvLIME, which can be accessed through the \texttt{predict\_parts()} function with specified \texttt{type} argument.

\paragraph{SurvSHAP(t).}
The default \texttt{type} in the \texttt{predict\_parts()} function is \texttt{'survshap'}, which refers to the SurvSHAP(t) method \citep{survshap}. This technique is inspired by the SHAP method \citep{shap} and adopts a game theoretical approach to estimate the contribution of each variable to the prediction of the model for a selected observation. The method was the first, specifically proposed for explaining survival models that provides so-called \emph{time-dependent} explainability. It does so via estimating variable attributions at every selected time point, i.e., for the selected variable $j$ and time point $t$, the technique estimates $\phi_t(\bx, j)$. 

The attributions can be calculated using various algorithms, including the KernelSHAP algorithm available through the \textbf{kernelshap} package \citep{kernelshap-package} or as an internal implementation, and the TreeSHAP algorithm \citep{treeshap} utilizing the \textbf{treeshap} package \citep{treeshap-package}. 

The calculated attributions are accurate, which means that after incorporating the mean prediction $\phi_{t}^0 = \frac{1}{n}\sum_{i=1}^n f_t(\bx_i)$, they sum up to the model's prediction, i.e.
\begin{align}
    f_t(\bx) = \phi_{t}^0 + \sum_{i=1}^p \phi_{t}(\bx, i).
\end{align}

Furthermore, for each variable, the resulting SurvSHAP(t) function can be aggregated to determine the local importance. One way to do it is to average absolute values over time by integrating, i.e.,
\begin{align}
\psi(\bx, j) = \int_{t_0}^{t_k} |\phi_{t}(\bx, j)|  \, \mathrm{d}w(t),
\end{align}
where $w(t)$ represents a weight function, by default $w(t) = \frac{t}{t_k}$.

\paragraph{SurvLIME.} Another available \texttt{type} is \texttt{'survlime'}, which corresponds to the SurvLIME method \citep{survlime}, drawing inspiration from the LIME technique \citep{lime}. SurvLIME aims to explain the prediction of the black-box survival model by its local approximation using a simple, interpretable surrogate model. For this purpose, a Cox proportional hazards model \citep{cph} is used, and it operates on the set of synthetic observations $\bx_1^s, \bx_2^s, \ldots, \bx_{n_s}^s$ generated randomly within the neighborhood of the observation to be explained (using Gaussian noise). 

The variables' attributions are determined using the coefficients of the Cox model. The authors of the method proposed to use the raw coefficient vector $\psi(\bx) = [\psi(\bx, 1), \psi(\bx, 2), \ldots, \psi(\bx, p)]$, which is calculated in the following optimization problem:
\begin{align}
    \psi(\bx) = \argmin_{\mathbf{b}} \sum_{i=1}^{n} w_i \sum_{j=0}^k v_{i,j}^2 \left( \ln{f_{t_j}^\mathrm{CHF}(\bx_i)} - \ln{H_{t_j}^{0}}  -\mathbf{b}^{T}\bx_i \right)^2 \left( t_{j+1} - t_j\right),
\end{align}
where $w_i$ are weights related to the distance between observation $\bx_i^s$ and $\bx$, while $H^{0}$ is the baseline cumulative hazard function estimated on the generated neighbourhood for the surrogate Cox model. 

In \textbf{survex}, in order to make the attributions comparable between variables of different scales, the default option is to consider the coefficients multiplied by the values of the variables for the observation being explained, i.e., 
\begin{align}
    \Tilde{\psi}(\bx, j) = \psi(\bx, j) \cdot \bx^{j}.
\end{align}

Note that SurvLIME has a notable limitation related to the challenges associated with the notion of an observation's neighborhood. Both the generation of new observations and the calculation of their weights may pose difficulties.  
In the \textbf{survex} implementation, the default approach mirrors that of the \textbf{lime} Python package. It employs Gaussian sampling and an exponential kernel with a default width set at $b=0.75\sqrt{p}$. There is also an alternative option, suggested by \citeauthor{pachon2023survlimepy} and used in the \textbf{SurvLIMEpy} package. This alternative method draws inspiration from non-parametric density estimation theory and employs a kernel width of $b=(\frac{4}{n(p+2)})^{\frac{1}{p+4}}$.

\subsection{Variable dependence}

Local variable dependence methods aim to understand the relationship between different variables and the model's prediction for a selected observation, focusing on the effects of a change in the variable on the outcome. For this purpose, \textbf{survex} implements individual conditional expectation curves (ICE curves), also known as Ceteris Paribus profiles (CP profiles),  in the \texttt{predict\_profile()} function.

The \texttt{predict\_profile()} function calculates ICE profiles for a specific observation as proposed by \cite{ice}. Minor adjustments to the fundamental method are requisite to incorporate the temporal dimension present in the survival setting. The ICE profile function describes the dependence of the (approximated) conditional expected value (prediction) of the outcome on some realization $z$ of variable $j$, using the prediction function $f$ at a specific time point $t$, that is

\begin{align}
    \mathrm{ICE}_t(f, \bx, j, z) =  f_t(\bx^{j|=z}).
\end{align}

Note that, $\bx^{j|=z}$ denotes the observation $\bx$ with the realization of variable $j$ set to value $z$, which assumes values over the entire observed range for the variable. In \textbf{survex}, by default $101$ uniform grid points are sampled within the range of $[\min(\bx^j),\max(\bx^j)]$. All other explanatory variables retain their fixed values as specified by the observation of interest $\bx$. Distinct ICE curves are obtained for different time points $t$. 

\section{Global explanation methods}
\label{sec:global}

\subsection{Measuring performance}

\textbf{survex} enables to assess models' predictive capabilities using the \texttt{model\_performance()} function, which calculates different metrics specific for survival models. By default, two different time-dependent metrics are calculated, along with their integrated versions and the widely used concordance index.

The first time-dependent metric is the Brier score \citep{brier, graf}, which for the time point $t$ is calculated as
\begin{align}
    \mathrm{BS}_t(f, \bX) =  I(y_i > t)
        \frac{(1 - f_t^{\mathrm{SF}}(\bx_i))^2}{\hat{G}(t)} + \frac{1}{n} \sum_{i=1}^n I(y_i \leq t \land \delta_i = 1)
        \frac{(0 - f_t^{\mathrm{SF}}(\bx_i))^2}{\hat{G}(y_i)},
\end{align}
where $1/\hat{G}(t)$ is the inverse probability of censoring weight estimated using Kaplan-Meier estimator. It can be understood as the mean squared error at the time point $t$ with incorporated censoring information. The true value for the selected observation is 1 if the observed time is greater than $t$ while it is equal to 0 if the observed time is lower or equal $t$ and the event occurred. For this metric, lower scores are better.

Another implemented time-dependent metric is the cumulative/dynamic area under receiver operating characteristics curve (C/D AUC). At the time point $t$, it is calculated as 
\begin{align}
    \mathrm{AUC}^{c/d}_t(f, \bX) =
        \frac{\sum_{i=1}^n \sum_{j=1}^n I(y_j > t) I(y_i \leq t) \omega_i
        I(f_t^{\mathrm{SF}}(\mathbf{x}_j) \leq f_t^{\mathrm{SF}}(\mathbf{x}_i))}
        {(\sum_{i=1}^n I(y_i > t)) (\sum_{i=1}^n I(y_i \leq t) \omega_i)},
\end{align}
where $\omega_i = 1/\hat{G}(t)$. It measures how well the model of interest differentiate observations who experienced the event of interest by a time $t$ (cumulative cases) from observations for which the event occurred after this time (dynamic controls). The higher the C/D AUC, the better. 

Both time-dependent metrics can be also aggregated using integration over time from $t_0$ to $t_k$ with a weight function $w(t)$ to obtain a single-number metric. By default $w(t) = \frac{t}{t_k}$. Another metric is the concordance index (C-index) \citep{cindex}, which is not time-dependent and utilizes the prediction in the form of risk score. It quantifies ranking of risk scores produced by the model of interest using the following formula:
\begin{align}
    \mathrm{C}(f, \bX) = \frac{\sum_{i\not=j} I(f^{\mathrm{risk}}(\bx_i) < f^{\mathrm{risk}}(\bx_j) )I(y_i > y_j)\delta_j}{\sum_{i\not=j}I(y_i > y_j)\delta_j}.
\end{align} 
Note that only comparable pairs are considered and ties are not comparable. The higher values of C-index are better. 

Additionally, the user can provide their own metric, which is facilitated by auxiliary functions enabling to adapt loss functions from \textbf{mlr3proba} (\texttt{loss\_adapt\_mlr3proba()}) and calculating an integral with respect to a selected weight function (\texttt{loss\_integrate()}).

Moreover, the \texttt{model\_performance()} function with \texttt{type} set to \texttt{'rocs'} makes it possible to create Receiver Operating Characteristic (ROC) curves for selected time points, by treating the values of the survival function at that time as probabilities of survival.  

\subsection{Diagnostics}
The fit of the model of interest can also be evaluated by diagnostic analysis of its residuals. In \textbf{survex}, the \texttt{model\_diagnostics()} function calculates multiple types of survival residuals, providing insights into the model's predictions and helping users diagnose any issues. 

The first type of residuals computed are Cox-Snell residuals \citep{coxsnell}, which rely on predicted cumulative hazard functions. Specifically, for an observation $\bx_i$, the corresponding Cox-Snell residual is defined as the cumulative hazard function's value at the observed time point $y_i$, i.e.,
\begin{align}
    r^C_i = f^{CHF}_{y_i}(\bx_i).
\end{align}
These residuals have the property that if the model fits the data, residuals follow an exponential distribution with a rate parameter of $\lambda = 1$. Consequently, diagnostic analysis involves comparing the cumulative hazard estimator for residuals with that of the standard exponential distribution. 

Cox-Snell residuals serve as the foundation for calculating the next type of residuals, known as martingale residuals \citep{martingale_deviance}, which are defined as 
\begin{align}
    r^M_i = \delta_i - r^C_i. 
\end{align}
Positive values indicate that the event of interest occurred sooner than expected based on the model's prediction, while negative values imply the opposite. 

Lastly, deviance residuals \citep{martingale_deviance} are estimated using martingale ones. Specifically, they are calculated as 
\begin{align}
    r^D_i = \mathrm{sgn}(r^M_i) \sqrt{-2\{r^M_i + \delta_i \ln (\delta_i - r^M_i) \}},
\end{align}
which makes them centered around 0 and, thus, easier to interpret. 

Martingale and deviance residuals can be plotted against observed times, variable values, or each other. Analyzing these residuals may help detect outliers or functional dependencies on variables.

\subsection{Variable importance}
Variable importance methods allow to understand the time-dependent importance of individual variables for a survival model in a global context (not only for one prediction). Importance scores can be calculated in \textbf{survex} using the \texttt{model\_parts()} function. It utilizes the permutation variable importance concept, i.e., describes the importance of $j$-th variable as a change in the loss function $\mathcal{L}$ caused by permutations of this variable in the dataset, i.e., 
\begin{align}
    \mathrm{PFI}_t(f, \bX, j, \mathcal{L}, \by) = \frac{1}{B} \sum_{i=1}^B \left( \mathcal{L}(f, \bX, \by) - \mathcal{L}(f, \bX^{*j_i}, \by)\right).
\end{align}
Here, $\mathcal{L}$ represents the loss function chosen to evaluate model performance, $\bX^{*j_i}$ denotes the $i$-th permutation of variable $j$ within the dataset $\bX$, and $B$ is the number of different permutations. Permuting a variable is supposed to simulate the loss of information associated with the variable.

By default, the Brier score is used as a loss function, but users can select other implemented metrics or provide their custom function to calculate the loss. If the chosen loss function returns a single value, as in the case of integrated metrics or C-index, then the variable importance scores are also not time-dependent. 

Another option for the assessment of time-dependent variable importance is to use the global aggregated version of SurvSHAP(t). After calculating SurvSHAP(t) results for every considered observation, the absolute values obtained can be averaged. \textbf{survex} allows such a procedure through \texttt{model\_survshap()} function. Furthermore, the results can be leveraged to analyze the distribution of individual variables' attributions concerning variable values, providing additional insights beyond importance scores alone. 

\subsection{Variable dependence}

Global variable dependence methods provide a comprehensive view of how different variables affect the model's predictions for all observations. The \textbf{survex} package provides access to two distinct techniques -- partial dependence plots and accumulated local effects plots -- by utilizing the \texttt{model\_profile()} and the \texttt{model\_profile\_2d()} functions with the appropriate type argument specification. 

\textbf{Partial Dependence.} The default type in the \texttt{model\_profile()} function is \texttt{'partial'}, which induces the computation of partial dependence (PD) profiles or plots with minor adjustments to the original method by \cite{pdp} for the inclusion of the time dimension present in the survival setting. The partial dependence function describes how the expected value of the model prediction varies with respect to a chosen explanatory variable, over a set of different, individually fixed grid points $z$, while all other variables fluctuate following their respective marginal distributions. In practice, a one-dimensional PD profile is estimated by the mean of the ICE profiles for all $n$ observations $\bx \in \bX$ from a dataset, that is

\begin{align}
    \mathrm{PDP}_t(f, \bX, j, z) = \frac{1}{n} \sum_{\bx \in \bX} f_t(\bx^{j|=z}).
\end{align}

Different PD profiles are obtained for different time points $t$. The \texttt{model\_profile\_2d()} generates two-dimensional profiles for two variables, by computing the average of ICE profiles obtained by varying two variables $j$ and $k$ over grid values within their respective observed range, while keeping all other variables fixed at their observed values for each specific observation. 

\textbf{Accumulated Local Effects.} The second available type is \texttt{'accumulated'}, which refers to accumulated local (AL) profiles or effects, as introduced by \cite{ale}. Again minor adjustments are made to account for different temporal dimensions. Equivalent to PD profiles, AL profiles aim to quantify and visualize how changes in a specific variable affect model predictions over all observations; however, these two approaches diverge in their estimation methodologies. Instead of considering the average impact of a Ceteris Paribus change in a singular variable, AL profiles capture and accumulate the local effects of variations in that variable. This distinction is particularly significant in addressing the extrapolation issue inherent in PD profiles. It arises when predictions or inferences outside the range of the observed data are made, particularly in the presence of strongly correlated variables, potentially leading to unreliable variable effect estimates. The \texttt{model\_profile()} function uses the estimation procedure proposed by \cite{ale}. Let $N_j(k) = \left( z_{k-1}^j,z_k^j\right]$ $(k = 1,\dots,K)$ be a partition of the range of the observed values $\bx^{j}$ of $\bX$ into $K$ intervals, with $z_0^j$ just below $\min(x_1^j,\dots,x_N^j)$ and $z_K^j = \max(x_1^j,\dots,x_N^j)$. Furthermore, $n_j(k)$ denotes the number of observations $\bx_i^{j}$ falling into $N_j(k)$, with $\sum_{k=1}^K n_j(k) = n$ and $k_j(z)$ is the index of interval $N_j(k)$ in which $z$ falls, then

\begin{align}
    \mathrm{ALE}_t(f, \bX, j, z) = \sum_{k=1}^{k_j(z)}\frac{1}{n_j(k)} \sum_{i:x_i^j \in N_j(k)} \left\{f_t(\bx^{j|=z_k^j}) - f_t(\bx^{j|=z_{k-1}^j}) \right\}.
\end{align}

Finite differences within ICE profiles, computed at the interval boundaries of distinct intervals $N_j(k)$ for a chosen observation, provide estimations of the local effects. These are then averaged across all observations where the observed value of $\bX^j$ falls within the respective interval, and subsequently, they are cumulatively accumulated over preceding intervals. Note, that in \textbf{survex} by default AL profiles are not centered, thus enabling direct comparisons with PD profiles; however, centered plots can be obtained by setting the \texttt{'center'} argument to \texttt{TRUE}. 

\bibliographystyle{abbrvnat}
\bibliography{reference}